\documentclass[conference]{IEEEtran}
\IEEEoverridecommandlockouts
\usepackage{cite}
\usepackage{amsmath,amssymb,amsfonts}
\usepackage{algorithmic}
\usepackage{adjustbox}
\usepackage{graphicx}
\usepackage{multirow}
\usepackage{textcomp}
\usepackage{lastpage}
\usepackage{ctable}
\usepackage{fancyhdr}

\newcommand{\rom}[1]{\uppercase\expandafter{\romannumeral #1\relax}}

\def\BibTeX{{\rm B\kern-.05em{\sc i\kern-.025em b}\kern-.08em
		T\kern-.1667em\lower.7ex\hbox{E}\kern-.125emX}}
\begin{document}
	
	\title{Dissimilarity-based representation for radiomics applications}
	\author{
		\IEEEauthorblockN{Hongliu CAO\IEEEauthorrefmark{1}\IEEEauthorrefmark{2}, Simon BERNARD\IEEEauthorrefmark{1}, Laurent HEUTTE\IEEEauthorrefmark{1} and Robert SABOURIN\IEEEauthorrefmark{2}}
		\IEEEauthorblockA{\IEEEauthorrefmark{1}Universit\'e de Rouen Normandie, LITIS (EA 4108), BP 12 - 76801 Saint-\'Etienne du Rouvray, France}
		\IEEEauthorblockA{\IEEEauthorrefmark{2}Laboratoire d'Imagerie, de Vision et d'Intelligence Artificielle, \'Ecole de Technologie Sup\'erieure, \\ Universit\'e du Qu\'ebec, Montreal, Canada}
	}
	
	
	
	\maketitle
	
	\begin{abstract}
		Radiomics is a term which refers to the analysis of the large amount of quantitative tumor features extracted from medical images to find useful predictive, diagnostic or prognostic information. Many recent studies have proved that radiomics can offer a lot of useful information that physicians cannot extract from the medical images and can be associated with other information like gene or protein data. However, most of the classification studies in radiomics report the use of feature selection methods without identifying the machine learning challenges behind radiomics. In this paper, we first show that the radiomics problem should be viewed as an high dimensional, low sample size, multi view learning problem, then we compare different solutions proposed in multi view learning for classifying radiomics data. Our experiments, conducted on several real world multi view datasets, show that the intermediate integration methods work significantly better than filter and embedded feature selection methods commonly used in radiomics. 
	\end{abstract}
	
	
	\section{Introduction}
	
	One of the biggest challenges of cancer treatment is that every tumor is different, known as tumor heterogeneity. It demands for more personalized treatment. The normal process of cancer detection is from certain signs and symptoms to the further investigation by medical imaging and at last confirmed by biopsy. However, with the improvement of medical imaging technology, tumor phenotype characteristics can be visualized in a non-invasive way \cite{coroller2015ct,aerts2014decoding}.

	Since 2012, by combining the word “radiology” and the suffix “omics”, a new term, \textit{radiomics}, was introduced, which refers to the process of extracting large amount of features from standard-of-care images obtained with CT (computed tomography), PET (positron emission tomography) or MRI (magnetic resonance imaging) to build descriptive, predictive or prognostic models for different cancers \cite{lambin2012radiomics}. Compared to the current qualitative analysis in radiology, radiomics can provide a quantitative analysis including much more useful information to make optimal treatment decisions and make cancer treatment more effective and less expensive \cite{kumar2012radiomics}. Many studies focus on the prediction of survival patients or the prediction of the response of patients to the treatment. A lot of classification tasks like classifying between patients with cancer and without cancer have also been done. Radiomics data can also be combined with genomics data and clinical data to improve the accuracy.
	
	From a machine learning point of view, radiomics is challenging in three ways:
	\begin{enumerate}
		\item \textbf{Small sample size}: Most of radiomics data sets have no more than 200 patients, many studies have even fewer than 100 patients. And the data sharing is very difficult because of different laws or politics issues.
		\item \textbf{High dimensional feature space}: As radiomics aims at extracting large amount of different features from medical images, the feature space is always high dimensional. Even though there is no quantitative definition of 'large amount', most of radiomics studies used at least 200 features. For example, in the work of \cite{aerts2014decoding,coroller2015ct}, they used over 400 features, and in the work of \cite{zhou2017mri}, they used 6746 features.
		\item \textbf{Multiple feature groups}: In order to obtain more useful information, multiple feature groups are extracted for radiomics data. These feature groups can come from different sources, different feature extractors or be of different natures, and each of them brings useful and complementary information, e.g. in many radiomics studies, the extracted features come from tumor intensity, tumor shape, texture of the tumor \cite{aerts2014decoding}. Exploiting the complementary information that different groups contain is a challenging task for improving the learning performance. 
	\end{enumerate}
	
	In most of the state-of-the-art works in radiomics, the multiple feature groups are concatenated all together in a very high dimension feature space, which results in an HDLSS (high dimension low sample size) machine learning task. Hence feature selection is a most commonly used method to reduce the dimension. However, if only a small subset of the features is chosen, certainly a lot of useful information will be lost and the heterogeneity of the tumor cannot be well represented. By concatenating all feature groups together into one single view of the patient, the complementary information that different feature groups can offer may be ignored.
	
	In contrast with the current studies that treat radiomics data as a single view machine learning problem, we propose in this paper to consider an HDLSS multi-view learning framework. Multi-View learning is a machine learning framework where data are represented by multiple distinct feature groups, and each feature group is referred to as a particular view\cite{xu2013survey}. We propose to compare the feature selection solutions commonly used in radiomics with classical multi-view learning solutions that are meant to  improve the performance by exploiting the information from different views of the same input data\cite{xu2013survey}. 
	
	The remainder of this paper is organized as follows: the related works in radiomics applications and multi-view learning are discussed in Section \rom{2}. In section \rom{3}, two dissimilarity based multi-view learning solutions are introduced.  Before turning to the result section (Section \rom{5}), we describe the data sets chosen in this study and provide the protocol of our experiments in Section \rom{4}. The final conclusion and future works are given in Section \rom{6}.
    
	\section{Related Works}
	This section firstly gives a brief overview of radiomics literature from a machine learning point of view. It secondly presents the multi-view learning approaches that could be straightforwardly applied to radiomics.
	\subsection{Machine learning methods used in radiomics}
    In a large majority of radiomics works, the multiple views are concatenated to form a single-view feature vector. As explained in the introduction, it usually leads to HDLSS issues. As a consequence, feature selection methods are systematically used to overcome the difficulty of learning in high dimension. 

Feature selection methods are traditionally divided into three main categories: filter, wrapper and embedded methods \cite{bolon2013review}. Radiomics works have mainly investigated filter selection methods. In general, MRMR (minimum redundancy maximum relevance), RELF (relief) and MIFS (mutual information feature selection) have better results than other filter methods in radiomics \cite{coroller2015ct,wu2016exploratory}. In the work of \cite{wu2016exploratory}, the authors investigated 24 feature selection methods and three classification methods for histology prediction: RELF showed higher prediction accuracy as compared to other methods in multivariate analysis. RELF was also compared with wrapper methods for classification of progression free survival in the work of \cite{farhidzadeh2016classification}, and features selected by RELF had better results. For other HDLSS problems, RELF also achieved very good performance\cite{bolon2013review}.
    
    In addition to those filter approaches, a few radiomics works have applied a successful embedded feature selection method, namely SVMRFE (support vector machine recursive feature elimination)  \cite{guyon2002gene}. This approach differs from the filter approaches used in radiomics by embedding the feature selection into the learning procedure, so that it can take the resulting classifier performance into account. In the work of \cite{Wang2017,zhang2017radiomics}, the authors showed that SVMRFE had very good performance on radiomics data. For other HDLSS data, SVMRFE also had very good performance in the work of \cite{bolon2013review}.
    
	\subsection{Multi-view learning methods}
	Multi-view learning methods can be divided into three main categories: early, intermediate and late integration approaches \cite{serra2015mvda}. Early integration refers to the approach discussed in the previous subsection, that is to say concatenating the multiple views and tackling the learning task from the resulting single-view feature space (e.g. with feature selection, and learning afterwards).
	
    Late integration consists in learning a different model separately on each view and aggregating the decisions. The most popular approach in this category is the Co-training method. It is a semi-supervised method that maximizes the mutual agreement on two distinct views, by exploiting the outcomes of two classifiers, one per view, on a subset of unlabeled data \cite{xu2013survey}. As explained in the introduction section, radiomics datasets are often made up of very few labeled instances and no additional unlabeled instances are available. As a consequence, co-training approaches are not straightforwardly applicable to radiomics learning tasks. 
Another method for late integration is to use a classical Multiple Classifier System approach (MCS). MCS combines the outputs of different classifiers trained on each view separately to improve total performance. In the work of \cite{tuarob2014ensemble}, the authors proposed to use five heterogeneous feature groups which represent different aspects of semantics for identifying health related messages in social media. Then they chose five classifiers for each feature group, and used MCS methods to combine the results together. Simply integrating the results of different classifiers makes MCS fast, efficient and flexible. 
    
	Finally, the third category in multi-view learning literature, namely the intermediate integration, is an in-between approach that builds an intermediate representation of the views, for better combining them before learning. A typical example is the dissimilarity-based learning, that projects each view in a space in which the samples are described by their dissimilarities to all the training instances. In that way, each view is separately projected in the same description space. Further details about this principle are given in the next section. In the work of \cite{gray2013random}, the authors used multi-modalities data for Alzheimer's Disease patients. To combine four modalities together, they calculated a Random forest similarity matrix for each modality and then fused the four dissimilarity matrices by averaging. Finally multidimensional scaling was used on the joint dissimilarity matrix, followed by a random forest classifier in the embedded dissimilarity space.	
    
	\subsection{Discussion}
	Most of the studies in radiomics have made use of filter selection methods, because they are independent from the classifier, simple to implement and computationally fast. But they may also easily filter some useful information for the classification task, whereas the objective of extracting a large number of features is exactly to bring additional information: if only a small subset of the features is chosen, certainly a lot of useful information is lost and the heterogeneity cannot be well represented. Filter methods also ignore the interaction with the classifier, the search in the feature subset space is separated from the search in the hypothesis space\cite{xu2013survey}. 
	
	In contrast to early integration, the two other multi-view learning methods, intermediate and late integration, can deal with feature heterogeneity by introducing one function to model a particular view and jointly optimize all the functions to improve the learning performance \cite{xu2013survey}. The late integration methods focus more on finding the agreement among views, whereas the intermediate methods use the same idea as in data fusion to generate a better representation of data by taking advantage of complementary information contained in each view. Intermediate integration seems thus more appropriate for radiomics data. 
	
	\section{ DISSIMILARITY-BASED LEARNING}
	As discussed above, intermediate integration methods can generate a better representation of data by taking advantage of the complementary information contained in each view. However, the question of how to integrate information coming from different views is a challenge because different views may have different feature types, feature sizes, and are not directly comparable. Projecting each view of the data in some dissimilarity space can offer a smart solution to that issue as views become comparable from one dissimilarity space to another (same feature type, same feature size) and the dimension of the initial HDLSS data is reduced.
    
\textbf{Dissimilarity matrix}: Let $\textbf{T}$ = $\{(\mathbf{X}_1, y_1), (\mathbf{X}_2, y_2), \dots, (\mathbf{X}_N, y_N)\}$ denote a training set made up of $N$ instances $\mathbf{X}_i$ from a domain $\mathbb{X}$, each one labeled with its true class $y_i$. A dissimilarity measure $d$ is a function from $\mathbb{X}^2$ to $\mathbb{R^+}$ that estimates how dissimilar two instances are. For two given instances $\mathbf{X}_i$ and $\mathbf{X}_j$, a high value $d(\mathbf{X}_i,\mathbf{X}_j)$ means that the two instances are very "different", while on the opposite, a low $d(\mathbf{X}_i,\mathbf{X}_j)$ means they are very similar; in particular, $d(\mathbf{X}_i, \mathbf{X}_i) = 0$.

For classification problems, the dissimilarity between two instances from the same class is expected to be small, while on the contrary the dissimilarity between two instances from two different classes is expected to be high.
	
Now, let $\mathbf{D}$ denote a $N \times N$ dissimilarity matrix, built from a given dissimilarity measure $d$ and from a training set $\mathbf{T}$, as defined in Equation \eqref{matrix}:
\begin{equation}\label{matrix}
\textbf{D} =
	\begin{bmatrix}
		d_{11} & d_{12} & d_{13} & \dots  & d_{1N} \\
		d_{21} & d_{22} & d_{23} & \dots  & d_{2N} \\
		\vdots & \vdots & \vdots & \ddots & \vdots \\
		d_{N1} & d_{N2} & d_{N3} & \dots  & d_{NN}
	\end{bmatrix}
\end{equation}
where $d_{ij}$ denotes $d(\mathbf{X}_i, \mathbf{X}_j)$, for all $\mathbf{X}_i,\mathbf{X}_j \in \mathbf{T}$.
	
$\mathbf{D}$ is non-negative and respects the reflexivity condition. Such a dissimilarity matrix can be viewed as a new training set, where each training instance $\mathbf{X}_i$ is described by a vector $(d_{i1},d_{i2},\dots,d_{iN})$. In the same way, using its dissimilarity to each of the training instances, any new instance $\mathbf{X}$ can be mapped into a $N$ dimensional dissimilarity space $DS$. For HDLSS data, the dimension of this dissimilarity space is necessarily smaller than the dimension of the original feature space.
	
Typically, a distance measure such as the euclidean distance can be used to measure dissimilarities. However, such a measure does not capture the class membership, which is an important criterion for classification tasks to tell whether or not two instances are similar. Compared to such a distance function without class information, class based dissimilarity measures are more powerful, e.g. the Random Forest dissimilarity measure \cite{shi2012unsupervised}.

\textbf{Random Forest}: Given a training set $\mathbf{T}$, a Random Forest classifier $\mathbf{H}$ is a classifier made up of $M$ trees, denoted as in Equation \eqref{e2}:
	\begin{equation}\label{e2}
	\mathbf{H}(\mathbf{X}) = \{h_k(\mathbf{X}),k=1,\dots,M\}
	\end{equation}
where $h_k(\mathbf{X})$ is a random tree grown using bagging and the random feature selection \cite{biau2016random}. For predicting the class of any new instance $\mathbf{X}$ with such a tree, $\mathbf{X}$ goes down the tree structure, from its root till one of its terminal nodes (or leaves). The descending path is decided by successive tests on the values of the features of $\mathbf{X}$, one per node. The prediction is given by the terminal node in which $\mathbf{X}$ falls. We refer the reader to \cite{biau2016random} for more information about this process.

Hence if two different instances fall in the same terminal node, they are likely to belong to the same class and they are also likely to share similarities between features, since they have followed the same descending path. 
	
	\textbf{Random Forest Dissimilarity (RFD)}: the RFD measure is inferred from a RF classifier $\mathbf{H}$, trained from $\mathbf{T}$. Let firstly define a dissimilarity measure $d^{(k)}$ inferred by the $k$th decision tree ${h_k}$: let $L_k$ denote the set of leaves of ${h_k}$, and let $l_k(\mathbf{X})$ denote a function from $\mathbb{X}$ to $L_k$ that returns the leaf node of ${h_k}$ where a given instance $\mathbf{X}$ falls when one wants to predict its class. The dissimilarity measure $d^{(k)}$, inferred by ${h_k}$, is defined as in Equation \eqref{sk}: if two training instances $\mathbf{X}_i$ and $\mathbf{X}_j$ fall in the same leaf of ${h_k}$, then the dissimilarity between both instances is set to 0, else to 1. 
	\begin{equation}\label{sk}
	d^{(k)}(\mathbf{X}_i, \mathbf{X}_j)=
	\begin{cases}
	0, & \text{if}\ l_k(\mathbf{X}_i) = l_k(\mathbf{X}_j)\\
	1, & \text{otherwise}
	\end{cases}
	\end{equation}

The RFD measure $d^{(\mathbf{H})}$ consists in calculating the $d^{(k)}$ value for each tree ${h_k}$ in the forest, and to average the resulting dissimilarity values over the $M$ trees, as in Equation \eqref{simil}:
	\begin{equation}\label{simil}
	d^{(\mathbf{H})}(\mathbf{X}_i, \mathbf{X}_j) = \frac{1}{M}\sum_{k=1}^{M} d^{(k)}(\mathbf{X}_i, \mathbf{X}_j)
	\end{equation}
	
\textbf{Multi-view learning dissimilarities}: For multi-view learning tasks, the training set $\mathbf{T}$ is composed of $Q$ views: $\mathbf{T}^{(q)} = \{(\mathbf{X}_1^{(q)}, y_1),\dots,(\mathbf{X}_N^{(q)}, y_N)\}$, q=1..Q. Firstly, for each view $\mathbf{T}^{(q)}$, the RFD matrices are computed as in Equation \ref{matrix}, and noted $\{\mathbf{D}_{\mathbf{H}}^{(q)}, q =1..Q \}$. In multi-view learning, the joint dissimilarity matrix $\mathbf{D}_{\mathbf{H}}$ can typically be computed by averaging over these matrices (cf. Equation \eqref{av}). 
	
\begin{equation}\label{av}
\mathbf{D}_{\mathbf{H}} = \frac{1}{Q}\sum_{q=1}^{Q}\mathbf{D}_{\mathbf{H}}^{(q)}
\end{equation}

For multi-view learning, this joint dissimilarity matrix $\mathbf{D}_{\mathbf{H}}$ can be used in two ways, either by using $\mathbf{D}_{\mathbf{H}}$ as a kernel matrix (denoted RFSVM method) or by using $\mathbf{D}_{\mathbf{H}}$ as a new training set (denoted RFDIS method):
\begin{enumerate}
		\item \textbf{Multi-view Random Forest kernel SVM (RFSVM)}: 
From the joint RFD matrix $\mathbf{D}_{\mathbf{H}}$ of Equation \eqref{av}, one can calculate the joint similarity matrix $\mathbf{S}_{\mathbf{H}}$ as
$\mathbf{S}_{\mathbf{H}} = \mathbf{1}- \mathbf{D}_{\mathbf{H}}$ where $\mathbf{1}$ is a matrix of ones. For SVM classifier, apart from the most used Gaussian kernel, user defined kernel matrix is also popular. Many studies have been done on user defined kernel matrix. For example, in the work of \cite{haasdonk2004learning}, they proposed to use the problem specific distance measure to construct a substitution Gaussian kernel. Similar to the idea in the work of \cite{englund2012novel}, the joint similarity matrix $\mathbf{S}_{\mathbf{H}}$ inferred from the RF classifier $\mathbf{H}$ is then used as a kernel matrix in a SVM classifier.
\item \textbf{Multi view random forest dissimilarity (RFDIS)}: RFDIS consists in learning a RF classifier $\mathbf{H}$ as if $\mathbf{D}_{\mathbf{H}}$ was a new training set. It is similar to the method described in \cite{gray2013random}. The joint dissimilarity vector is seen as a feature vector, and a random forest classifier is built on these new features.
\end{enumerate}

	\section{EXPERIMENTS}
	
	\subsection{Description of the data sets}
	In this study, we use several publicly available HDLSS multi-view datasets. A general information of all datasets can be found in Table \ref{tab:data}. The first four datasets are radiomics data. There are five views for each of these four datasets: four texture feature groups from axial T1-weighted MR images before and after gadolinium-based CE material administration as well as axial T2-weighted and axial T2-weighted fluid attenuated inversion recovery (FLAIR)  images; the fifth view is made up of vasari features. More details about this dataset can be found in the work of \cite{zhou2017mri}. LSVT is a dataset on vocal performance degradation of Parkinson's disease subjects with four groups of features extracted:  physiological observation, signal-to-noise ratio measure, wavelet measure and Mel frequency cepstral coefficients \cite{tsanas2014objective}. Metabolomic contains  biomarkers (CEA and TIMP), fluorescence concentration (PF) and NMR profiles for early detection of colorectal cancer \cite{bro2013data}. Cal7 \cite{li2015large}, Cal20 \cite{li2015large} and Mfeat \cite{frank2010uci} are image classification data using different feature extractors (600 instances are used here for Mfeat). BBC and BBCSport are text classification data constructed from the news article corpora by splitting articles into related segments of text \cite{xia2014robust}. 
	
	\begin{table}[htbp]
		\caption{\label{tab:data}The overview of each dataset. }	
		\begin{center}
			\begin{adjustbox}{max width=0.5\textwidth}
				\begin{tabular}{|c|ccccc|}
					\hline
					& \textbf{\#features} & \textbf{\#samples} &\textbf{\#views} &\textbf{\#classes}& \textbf{balanced class}  \\
					\hline
					nonIDH1\cite{zhou2017mri} & 6746 &84 & 5 & 2&no \\
					\hline
					IDHcodel\cite{zhou2017mri} & 6746 &67 & 5 & 2&no \\
					\hline
					lowGrade\cite{zhou2017mri} & 6746 &75 & 5 & 2&no \\
					\hline
					progression\cite{zhou2017mri} & 6746 &84 & 5 & 2&no \\
					\specialrule{.2em}{.1em}{.1em} 
					LSVT\cite{tsanas2014objective} & 309 &126 & 4 & 2&no \\
					\hline
					Metabolomic\cite{bro2013data} & 476 &94 & 3 & 2& yes \\
					\hline
					Cal7\cite{li2015large} & 3766 &1474 & 6 & 7& no  \\
					\hline
					Cal20\cite{li2015large} & 3766 &2386 & 6 & 20& no  \\
					\hline
					Mfeat\cite{frank2010uci} & 649 &600 & 6 & 10&yes \\
					\hline
					BBC\cite{xia2014robust} & 13628 &2012 &2 & 5& no  \\
					\hline
					BBCSport\cite{xia2014robust} & 6386 &544 & 2 & 5& no  \\
					\hline
				\end{tabular}
			\end{adjustbox}
		\end{center}
	\end{table}
	
	\subsection{Protocol of experiments}

	The main objective of our experiments is to show that muti-views approaches, in particular the two dissimilarity based intermediate integration approaches described in the previous section, are better alternatives for classifying radiomics data than the commonly used feature selection methods. For that purpose, six methods are compared: RELF and SVMRFE, the two feature selection methods mostly used in radiomics; RFSVM and RFDIS, the two intermediate integration methods described above; LateRF and LateRFDIS, two late integration methods. LateRF method builds a random forest classifier $\textbf{H}^{(q)}$ for each view of the data $\textbf{T}^{(q)}$, and then combines the results together by majority voting. LateRFDIS method firstly creates a RF dissimilarity space $DS^{(q)}$ for each view $q$, then builds a random forest classifier in each $DS^{(q)}$, and finally combines the results together by majority vote. Here random forest has been chosen because it can deal well with different data types and mixed variables, which can avoid searching for the best classifiers according to data types, data size or data complexity for each view \cite{shi2012unsupervised}. 
	
For the two feature selection methods, it is necessary to fix the number of selected features. However it is very difficult to find the best threshold, most authors just take the top-1 or top-5 features. Following the experiments of \cite{bolon2013review}, the rules we use to fix the number of selected features from the total number $n$ of initial features are the following: 
	\renewcommand{\theenumi}{\roman{enumi}}
	\begin{enumerate}
		\item if $n$ \textless 10, then select 75 \% of features
		\item if 10 \textless $n$ \textless 75, then select 40 \% of features
		\item if 75 \textless $n$ \textless 100, then select 10 \% of features
		\item if 100 \textless $n$ \textless 1000, then select 3 \% of features
		\item if $n$ \textgreater 1000, then select 25 features 
	\end{enumerate}
	
	For both feature selection methods, a random forest classifier is then built with selected features.

	For RFSVM, the search range of parameter $C$ for SVM is [0.01, 0.1, 1, 10, 100, 1000].
	
	For all random forest classifiers, the number of trees is set to 500 while other parameters are set by default with scikit-learn package for python.

	Note also that in \cite{bill2014comparative}, the authors found that when dealing with HDLSS data, stratification of the sampling is central for obtaining minimal misclassification. In this work, a stratified repeated random sampling approach was used to achieve a robust estimate of the performance. The stratified random splitting procedure is repeated 10 times, with 50\% sampling rate in each subset. In order to compare the methods, the mean and standard deviations of accuracy were evaluated over 10 runs.

	\section{Results}
	
	\subsection{Results on non-radiomics data}
	
	\begin{table}[htbp]
		\caption{\label{tab:50}Experimental results with 50\% training data 50\% test data for non-radiomics data}
		
		\begin{center}
			\begin{adjustbox}{max width=0.5\textwidth}
				\begin{tabular}{ | l  | p{1cm}| p{1.cm}| p{1.cm}| p{1.cm}|p{1.cm}| p{1.cm}|}
					
					\hline
					
					Dataset & 
					RELF +RF & SVMRFE +RF & RFSVM  & RFDIS&LateRF&LateRFDIS\\ 
					\hline
					
					LSVT & 
					$81.11\%\pm 5.04$ &
					$\textbf{84.12}\%\pm 3.48$ 
					&$\textbf{84.12}\%\pm2.93$
					&
					$83.33\%\pm3.97$
					&
					$80.47\%\pm2.01$
					&
					$81.42\%\pm2.66$
					\\ 
					\hline
					
					Metabolomic & 
					$60.83\%\pm6.37$ &
					$63.54\%\pm7.53$ 
					&$\textbf{68.75}\%\pm5.10$
					&
					$67.71\%\pm5.12$
					&
					$64.58\%\pm7.63$
					&
					$64.37\%\pm6.55$
					\\ 
					\hline
					
					Cal7& 
					$90.18\%\pm 2.08$ &
					$94.57\%\pm0.86$
					&$\textbf{96.36}\%\pm0.47$
					&
					$95.21\%\pm0.67$
					&
					$91.66\%\pm0.46$
					&
					$93.98\%\pm0.73$
					\\ 
					\hline
					
					Cal20& 
					$75.01\%\pm0.94$ &
					$ 85.06\%\pm1.98$ 
					&$88.39\%\pm0.34$
					&
					$\textbf{89.12}\%\pm0.69$
					&
					$81.57\%\pm0.33$
					&
					$86.15\%\pm0.58$
					\\ 
					\hline
					Mfeat 
					& $89.13\%\pm2.49$
&
$91.13\%\pm4.12$
&
$\textbf{97.83}\%\pm0.95$
&
$97.56\%\pm0.99$
&
$96.86\%\pm1.11$
&
$96.56\%\pm1.26$
					\\
					\hline
					BBC&
					$59.92\%\pm1.89$
					&
					$74.19\%\pm1.76$
					&
					$\textbf{95.63}\%\pm0.39$
					&
					$92.82\%\pm0.67$
					&
					$92.12\%\pm0.62$
					&
					$88.88\%\pm0.50$
					\\ 
					\hline
					BBCSport&
					$56.33\%\pm4.08$
					&
					$78.05\%\pm2.88$
					&
					$\textbf{95.56}\%\pm0.81$
					&
					$81.75\%\pm2.70$
					&
					$88.13\%\pm1.78$
					&
					$81.61\%\pm1.96$
					\\ 
					\specialrule{.2em}{.1em}{.1em} 
					Average Rank&5.82& 4.23& \textbf{1.55} & 2.55&  3.23&  3.64
					\\
					\hline
				\end{tabular}
			\end{adjustbox}
		\end{center}
	\end{table}
	For the experiments, the results on the seven non-radiomics HDLSS multi-view datasets are firstly presented. For data splits with 50\% training data and 50\% test data, the results of mean and standard deviation of accuracy over 10 repetitions are shown in Table \ref{tab:50}. It can be seen that in general the intermediate method RFSVM performs the best among all the six methods, and is ranked first for six of the seven datasets. By looking at the average ranking, the two dissimilarity based intermediate integration methods are the best, while the two feature selection methods are the worst.

	\begin{figure}[htbp]
		\centerline{\includegraphics[width=0.5\textwidth]{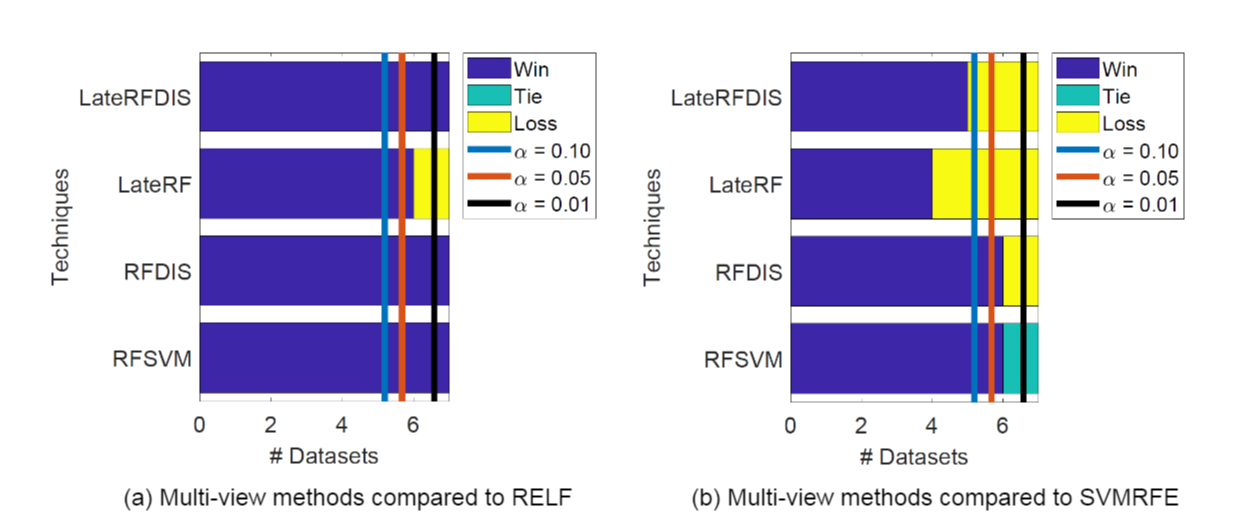}}
		\caption{\label{fig:posthoc2}Pairwise comparison between multi-view solutions and feature selection methods for non-radiomics data. The vertical lines illustrate the critical values considering a confidence level $\alpha$= \{0.10, 0.05, 0.01\}. }
	\end{figure}
	
	To see more clearly the difference, a pairwise analysis based on the Sign test is computed on the number of wins, ties and losses as in the work of \cite{cruz2018dynamic}. The result is shown in Figure \ref{fig:posthoc2}. The two intermediate and the two late integration methods are compared to RELF in Figure \ref{fig:posthoc2} (a) and compared to SVMRFE in Figure \ref{fig:posthoc2} (b). Each vertical line indicates the critical value corresponding to a confidence level $\alpha$. If the number of wins is greater than or equal to a critical value, it means that the corresponding method is significantly better than the baseline method. Figure \ref{fig:posthoc2} (a) shows that with $\alpha$ = 0.10 and 0.05, the four multi-view methods are significantly better than RELF. Figure \ref{fig:posthoc2} (b) shows that the two late integration methods are not always significantly better than SVMRFE, but the two intermediate integration methods are significantly better than SVMRFE with  $\alpha$ = 0.05. The results on the seven non-radiomics data confirm our hypothesis that dissimilarity based intermediate integration methods are significantly better than the state of art feature selection methods used in radiomics.

	\subsection{Results on radiomics data}
	\begin{table}[htbp]
		\caption{\label{tab:51}Experimental results with 50\% training data 50\% test data for radiomics data}
		
		\begin{center}
			\begin{adjustbox}{max width=0.5\textwidth}
				\begin{tabular}{ | l  | p{1cm}| p{1.cm}| p{1.cm}| p{1.cm}|p{1.cm}| p{1.cm}|}
					
					\hline
					
					Dataset & 
					RELF +RF & SVMRFE +RF & RFSVM  &RFDIS&LateRF&LateRFDIS\\ 
					\hline
					nonIDH1&
					
					$74.65\%\pm5.55$
					&
					$76.28\%\pm4.39$
					&
					$80.69\%\pm4.17$
					&
					$79.53\%\pm3.57$
					&
					$\textbf{82.79} \%\pm2.37$
					&
					$80.93\%\pm2.51$
					\\
					\hline
					IDHcodel&
					
					$72.94\%\pm4.89$
					&
					$73.23\%\pm5.50$
					&
					$\textbf{76.76}\%\pm4.52$
					&
					$76.47\%\pm3.95$
					&
					$\textbf{76.76}\%\pm2.06$
					&
					$76.17\%\pm2.06$
					\\
					\hline
					lowGrade&
					
					$60.46\%\pm5.79$
					&
					$62.55\%\pm3.36$
					&
					$63.95\%\pm4.56$
					&
					$63.48\%\pm3.76$
					&
					$64.41\%\pm3.76$
					&
					$\textbf{65.11}\%\pm5.20$
					\\
					\hline
					progression&
					$60.26\%\pm4.92$
					&
					$62.36\%\pm3.73$
					&
					$\textbf{65.52}\%\pm4.47$
					&
					$63.42\%\pm6.49$
					&
					$61.31\%\pm4.25$
					&
					$58.94\%\pm6.02$
					\\
					\specialrule{.2em}{.1em}{.1em} 
					Average Rank&5.750  &   4.50 &  \textbf{2.12}   &  3.25  &   \textbf{2.12}   &  3.25 
					\\
					\hline
				\end{tabular}
			\end{adjustbox}
		\end{center}
	\end{table}
	
	We now show that our hypothesis, validated on the seven non-radiomics datasets, can be confirmed on four real world radiomics datasets. The results of mean and standard deviation of accuracy over 10 repetitions are presented in Table \ref{tab:51}. The best performance on the four radiomics datasets are achieved two times by RFSVM, two times by LateRF and once by LateRFDIS. By looking at the average ranking, RFSVM and LateRF are ranked at the first place, while RFDIS and LateRFDIS are ranked at the second place.
	
	\begin{figure}[htbp]
		\centerline{\includegraphics[width=0.5\textwidth]{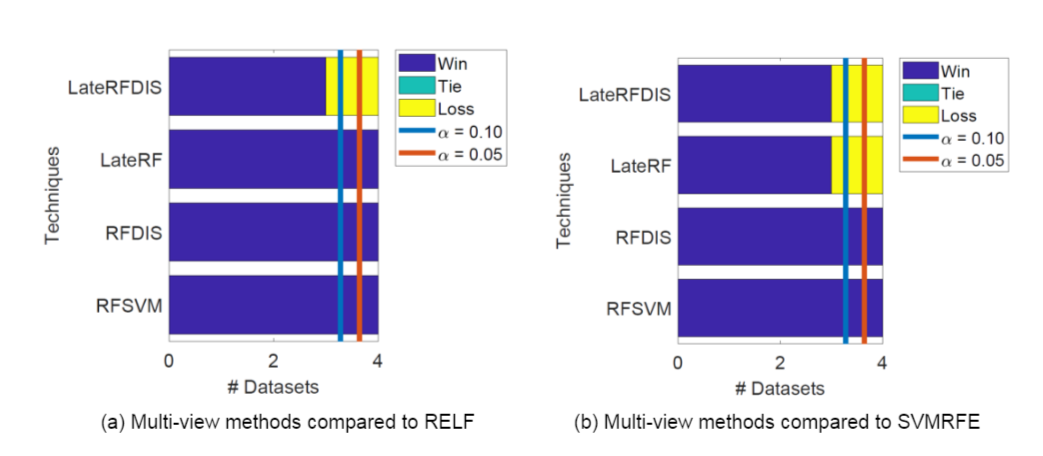}}
		\caption{\label{fig:posthoc3}Pairwise comparison between multi-view solutions and feature selection methods for radiomics data. The vertical lines illustrate the critical values considering a confidence level $\alpha$= \{0.10 , 0.05\}. The confidence level $\alpha$=0.01 is not shown due to the limit number of datasets.}
	\end{figure}
	
	Similar to the analysis for non-radiomics data, a pairwise analysis based on the Sign test is also done for the four radiomics datasets.  Figure \ref{fig:posthoc3} (a) shows that with $\alpha$ = 0.10 and 0.05, RFSVM, RFDIS and LateRF are significantly better than RELF, but LateRFDIS is not. Figure \ref{fig:posthoc3} (b) shows that the two late integration methods are not significantly better than SVMRFE, but the two intermediate integration methods are significantly better than SVMRFE with $\alpha$ = 0.10 and 0.05. 
	
	Note that these latter results on radiomics data may be explained by the fact that, even if radiomics and non-radiomics data used in these experiments are HDLSS, the ratio "feature size" over "sample size" is much bigger for radiomics data (100 times bigger) than  for non-radiomics data (1.5 to 12 times bigger). However, the dissimilarity based intermediate integration methods are proved to be significantly better than the two state of art early integration methods for both radiomics and non-radiomics data, which shows great potential for solving radiomics problem.

	\section{Conclusions}
	In this paper, we have tackled the problem of radiomics data classification as an HDLSS multi-view learning task. Contrary to the most commonly used methods in radiomics that concatenate the multiple feature groups into a single one view in an early integration manner and then select the best features for classification, we have shown that intermediate and late integration methods can offer better benefits for taking advantage of the complementary information brought by each view than the early integration methods.
    
	To confirm our hypothesis, we have compared two representative early integration methods (RELF and SVMRFE), two dissimilarity based intermediate integration methods (RFSVM and RFDIS) and two late integration methods (LateRF and LateRFDIS), across seven real-world non-radiomics datasets and four real-world radiomics datasets. Our experiments have shown that the two intermediate integration methods, RFSVM and RFDIS, are significantly better than the state-of-the-art early integration methods with RELF and SVMRFE. We have also shown that even if LateRF and LateRFDIS methods are  better than the two early integration methods with feature selection, they are not significantly better than SVMRFE. We can conclude that, for radiomics like data, the dissimilarity-based intermediate integration methods are a better alternative than the commonly used early integration methods. 
	
	As part of our future works, we aim at improving the quality of the dissimilarity space for each view by adapting the hyperparameters of the random forest based dissimilarity measure, as at present the same hyperparameters are used for each view. In this work, the two dissimilarity-based intermediate integration methods treat all the views equally, but a weighted combination could have also been used to generate a better joint dissimilarity matrix. Finally, none of the datasets we have tested in this work contain missing values and missing views. But in our future works we have to deal with that issue .

	\section*{Acknowledgment}
	This work is part of the DAISI project, co-financed by the European Union with the European Regional Development Fund (ERDF) and by the Normandy Region.
	
	\bibliographystyle{ieeetr}
	\bibliography{sample}
	
\end{document}